\documentclass[conference]{IEEEtran}
\IEEEoverridecommandlockouts

\usepackage{array}
\usepackage{multirow}
\usepackage{makecell}
\usepackage{booktabs}

\usepackage{cite}
\usepackage{amsmath,amssymb,amsfonts}
\usepackage{algorithmic}
\usepackage{graphicx}
\usepackage{textcomp}
\usepackage{xcolor}
\def\BibTeX{{\rm B\kern-.05em{\sc i\kern-.025em b}\kern-.08em
    T\kern-.1667em\lower.7ex\hbox{E}\kern-.125emX}}
\begin{document}
\bibliographystyle{unsrt} 

\title{Chat2Brain: A Method for Mapping Open-Ended Semantic Queries to Brain Activation Maps\\
\thanks{$^*$Corresponding author: Tuo Zhang (tuozhang@nwpu.edu.cn). Data used in the preparation of this article were obtained from the NeuroQuery database.}
}

\author{\IEEEauthorblockN{Yaonai Wei}
\IEEEauthorblockA{\textit{School of Automation} \\
\textit{Northwestern Polytechnical University}\\
Xi'an, China \\
rean$\_$wei@mail.nwpu.edu.cn}
\and
\IEEEauthorblockN{Tuo Zhang$^*$}
\IEEEauthorblockA{\textit{School of Automation} \\
\textit{Northwestern Polytechnical University}\\
Xi'an, China \\
tuozhang@nwpu.edu.cn}
\and
\IEEEauthorblockN{Han Zhang}
\IEEEauthorblockA{\textit{Network and Data Center} \\
\textit{Northwest university}\\
Xi'an, China \\
zhang$\_$han@stumail.nwu.edu.cn}
\and
\IEEEauthorblockN{Tianyang Zhong}
\IEEEauthorblockA{\textit{School of Automation} \\
\textit{Northwestern Polytechnical University}\\
Xi'an, China \\
2022100670@mail.nwpu.edu.cn}
\and
\IEEEauthorblockN{Lin Zhao}
\IEEEauthorblockA{\textit{School of Computing} \\
\textit{University of Georgia}\\
GA, USA \\
lin.zhao@uga.edu}
\and
\IEEEauthorblockN{Zhengliang Liu}
\IEEEauthorblockA{\textit{School of Computing} \\
\textit{University of Georgia}\\
GA, USA \\
zl18864@uga.edu}
\and
\IEEEauthorblockN{Chong Ma}
\IEEEauthorblockA{\textit{School of Automation} \\
\textit{Northwestern Polytechnical University}\\
Xi'an, China \\
mc-npu@mail.nwpu.edu.cn}
\and
\IEEEauthorblockN{Songyao Zhang}
\IEEEauthorblockA{\textit{School of Automation} \\
\textit{Northwestern Polytechnical University}\\
Xi'an, China \\
songyaozhang@mail.nwpu.edu.cn}
\and
\IEEEauthorblockN{Muheng Shang}
\IEEEauthorblockA{\textit{School of Automation} \\
\textit{Northwestern Polytechnical University}\\
Xi'an, China \\
shangmuheng@mail.nwpu.edu.cn}
\and
\IEEEauthorblockN{Lei Du}
\IEEEauthorblockA{\textit{School of Automation} \\
\textit{Northwestern Polytechnical University}\\
Xi'an, China \\
dulei@nwpu.edu.cn}
\and
\IEEEauthorblockN{Xiao Li}
\IEEEauthorblockA{\textit{School of Information Science$\&$Technology} \\
\textit{Northwest University}\\
Xi'an, China \\
lixiao@nwu.edu.cn}
\and
\IEEEauthorblockN{Tianming Liu}
\IEEEauthorblockA{\textit{School of Computing} \\
\textit{University of Georgia}\\
GA, USA \\
tianming.liu@gmail.com}
\and
\IEEEauthorblockN{Junwei Han}
\IEEEauthorblockA{\textit{School of Automation} \\
\textit{Northwestern Polytechnical University}\\
Xi'an, China \\
junweihan2010@gmail.com}
}

\maketitle

\begin{abstract}

Over decades, neuroscience has accumulated a wealth of research results in the text modality that can be used to explore cognitive processes. Meta-analysis is a typical method that successfully establishes a link from text queries to brain activation maps using these research results, but it still relies on an ideal query environment. In practical applications, text queries used for meta-analyses may encounter issues such as semantic redundancy and ambiguity, resulting in an inaccurate mapping to brain images. On the other hand, large language models (LLMs) like ChatGPT have shown great potential in tasks such as context understanding and reasoning, displaying a high degree of consistency with human natural language. Hence, LLMs could improve the connection between text modality and neuroscience, resolving existing challenges of meta-analyses. In this study, we propose a method called Chat2Brain that combines LLMs to basic text-2-image model, known as Text2Brain, to map open-ended semantic queries to brain activation maps in data-scarce and complex query environments. By utilizing the understanding and reasoning capabilities of LLMs, the performance of the mapping model is optimized by transferring text queries to semantic queries. We demonstrate that Chat2Brain can synthesize anatomically plausible neural activation patterns for more complex tasks of text queries.

\end{abstract}

\begin{IEEEkeywords}
meta-analysis, large language model, image generation
\end{IEEEkeywords}

\section{Introduction}

Neuroscience has accumulated a wealth of research findings over decades, continuously advancing our understanding of cognitive processes\cite{matsuo2010neural, king2014handyman}. To fully integrate and leverage these findings for new insights and be unconstrained by the number of subjects and cognitive processes that an individual find can investigate, numerous meta-analytic methods have been developed, bridging the modalities of text and brain activation\cite{gorgolewski2015neurovault, dockes2020neuroquery, ngo2021text2brain}. As so, researchers can explore potential cognitive processes of the brains, such as the activation of specific brain regions or an integrated brain network during certain tasks, through text queries. However, these methods are often idealized and impractical due to the semantic complexity of text queries in real-world scenarios, requiring guidance from domain experts\cite{costafreda2008predictors,minzenberg2009meta,shackman2011integration}. To be more specific, conventional methods of natural language processing struggle to address issues related to synonymy and complex semantics such as long text with redundancy and possible mistakes or ambiguous semantic information which needs powerful ability of context understanding and reasoning\cite{carp2012secret,button2013power}. For example, Neurosynth\cite{yarkoni2011large} and Neuroquery\cite{dockes2020neuroquery} synthesize brain activation maps using keyword searches and the similarities between these keywords over the text corpus. However, the vocabulary used in text queries may have different meanings in different contexts. Hence, keyword searches make it challenging to tackle synonymous queries. While Text2Brain\cite{ngo2021text2brain} establishes a method to generate brain activation maps from free-form text queries by leveraging the language model to some extent merge synonyms, it still falls short in dealing with more complex semantics limited by the scale of the language model.

Meanwhile, with recent advancements in hardware technology and algorithms, powerful large language models (LLMs) have emerged, demonstrating context understanding and reasoning capabilities that previous language models lack\cite{dai2023chataug, ma2023impressiongpt}. In particular, Ma et al. \cite{ma2023impressiongpt} show the ability of LLMs to extract key semantic information and optimize the representation of semantics based on given texts. Therefore, it is straightforward to utilize the powerful capabilities of LLMs to organize conventional text queries in meta-analyses, untangle contextual relationships, and extract crucial semantic information, thereby transforming them into semantic queries.

In this work, we propose a method called Chat2Brain, as shown in Fig.\ref{fig:overview}, which combines Text2Brain (Model) with ChatGPT, to establish a two-stage mapping from free-form text queries to semantic queries and then to brain activation maps. By using ChatGPT, Chat2Brain extracts the most important semantic queries from various text queries, enabling accurate predictions of brain activation maps even in complex query environments. This method facilitates the exploration of cognitive processes in the human brain in a simpler manner, allowing users to conduct queries using vague or imprecise descriptions in a more user-friendly manner, since ChatGPT helps summarize semantics. We conduct experiments on Chat2Brain using both standard and non-standard query environments to evaluate its performance under different conditions. Additionally, we compare the model's prediction results with those of Neuroquery and Text2Brain on the same dataset. Our results demonstrate that Chat2Brain maintains powerful generation capabilities even in complex query environments, producing brain activation maps that align more closely with expectations.


\section{Related Works}

\subsection{Meta-Analysis}
Meta-analysis, first proposed by Fox et al.\cite{fox2002mapping}, has developed various methods, among which the classic ones in recent years are NeuroVault\cite{gorgolewski2015neurovault}, NeuroQuery\cite{dockes2020neuroquery}, and Text2Brain\cite{ngo2021text2brain}.


NeuroVault is a web-based repository that makes it easy to deposit and share statistical maps, focusing on addressing issues such as limited sharing of research data\cite{poline2012data} and subjective interpretations of research results.  NeuroQuery is a meta-analytic tool that predicts the neural correlates of neuroscience concepts related to behavior or anatomy. It is fitted using supervised machine learning on 13459 full-text publications and assembles results from the literature into brain activation maps using arbitrary queries with words from the vocabulary of 7547 neuroscience terms\cite{newell1973you}. Both methods rely on keyword searching to map text queries to brain activation maps, resulting in the challenge of free-form text queries and merging synonyms\cite{poldrack2016brain}.


Text2Brain is a generative model that allows free-form text queries as the input, which is different from NeuroVault and NeuroQuery. It introduces a language model, SciBERT, that enables synonyms merging and many-to-one mappings between text queries and brain activation maps. However, it is hard for Text2Brain to accept text queries that may contain complex semantics\cite{ngo2021text2brain}.

Taken together, it is necessary to introduce some kind of powerful language model that can transfer text queries to semantic queries by extracting key semantic information in text queries\cite{turney2010frequency}, which brings the more accurate prediction of corresponding brain activation maps.

\begin{figure}
    \centering
    \includegraphics[width=1\linewidth]{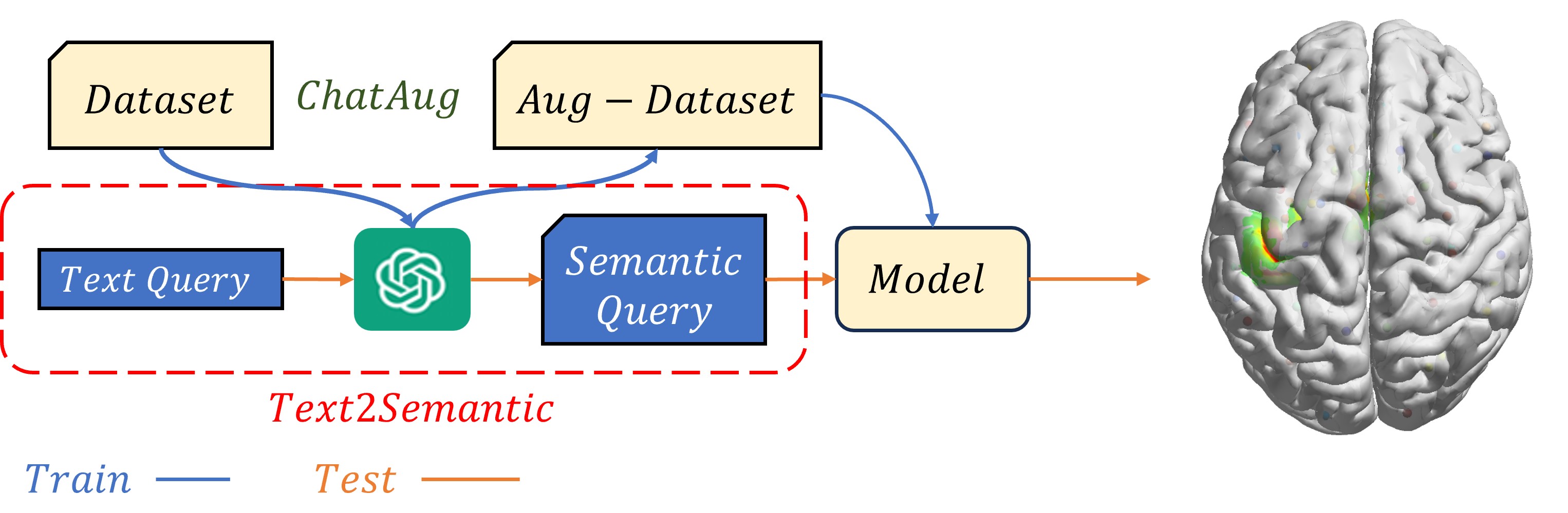}
    \caption{Overview of Chat2Brain}
    \label{fig:overview}
\end{figure}



\subsection{The Application of LLMs in Various Fields}

As one of the most influential artificial intelligence (AI) products today \cite{dai2023chataug}, LLMs known as generative models, provide a user-friendly human-machine interaction platform and have been rapidly integrated into various fields of application such as data augmentation, semantic extraction, reasoning, context understanding, and more \cite{dai2023chataug,liu2023summary,wang2023chatcad,ma2023impressiongpt,chowdhery2022palm,longpre2023flan,scao2022bloom,zhang2022opt}.

Recently, ImpressionGPT by Ma et al. \cite{ma2023impressiongpt} explored LLM's ability to comprehend radiology reports through an innovative dynamic prompting paradigm. The authors utilize LLMs' in-context learning ability by creating dynamic contexts with domain-specific, individualized sample findings-impression pairs. This approach enables the model to acquire contextual knowledge from semantically similar examples in existing data.


\begin{figure*}
    \centering
    \includegraphics[width=0.7\linewidth]{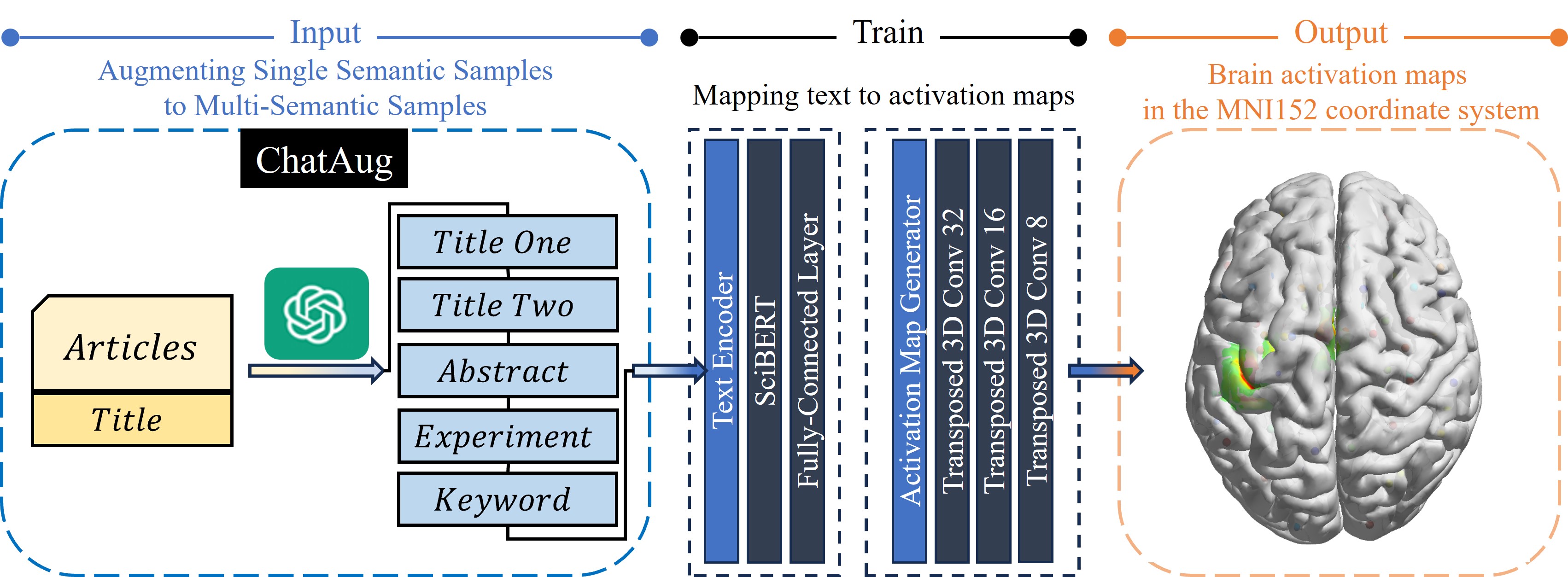}
    \caption{Flowchart of Chat2Brain during the training phase.}
    \label{fig:enter-label}
\end{figure*}

Li et al. \cite{dai2023auggpt} proposed a text data augmentation approach AugGPT, which rephrased each sentence in the training samples into multiple conceptually similar but semantically different samples. The augmented samples can then be used in downstream model training. Wu et al. \cite{wu2023exploring} compared the deductive reasoning abilities of LLMs and investigated ChatGPT and GPT-4's performance in the specialized domain of radiology, using a natural language inference task and comparing them to fine-tuned models. 

The various methods mentioned above have demonstrated the application value of LLMs in fields such as medicine and natural language processing, once again confirming their powerful abilities in context understanding, semantic extraction, and reasoning. This work represents one of the efforts to propose a framework that combines LLMs with neuroscience.


\section{Materials and Methods}

\subsection{Overview}

Fig.\ref{fig:overview} illustrates the overview of Chat2Brain, where the blue and yellow arrows represent the training and testing phases, respectively. During the training phase, the ChatAUG for data augmentation is used to help the model learn richer semantic connections between brain activations at a spatial location and key words in the text input. During the testing phase, dynamic semantic prompts are used to assist ChatGPT in transforming text queries into semantic queries which can be mapped to more accurate coordinates of brain activations due to the similarity between itself and key words learned by the model.

Chat2Brain can be decomposed into two modalities, three components, and two transformations. From the modality perspective, Chat2Brain encompasses the text modality as input and the brain activation map modality as output. From the methodological perspective, Chat2Brain employs ChatAUG as a data augmentation strategy (in Fig.\ref{fig:enter-label}), utilizes SciBERT\cite{beltagy2019scibert} as the text encoder, and employs the 3D convolutional layer as the brain activation map generator. From the transformation perspective, Chat2Brain encompasses two distinct transformations: text-to-semantics, which converts free-form text queries into semantic queries, and semantics-to-brain, which converts semantic queries into brain activation maps. Note that the modality is not changed during the transformation, where semantic queries are still as text modality but have more discriminating information related to the brain activation than text queries. Overall, Chat2Brain establishes a two-stage mapping model that progresses from free-form text queries to semantic queries and subsequently to brain activation maps.

The details of Chat2Brain including data preprocessing, model, ChatAUG, and Text2Semantics will be demonstrated as follows.

\subsection{Data Preprocessing}

The dataset used in this study is based on 13,460 research articles related to neuroscience, publicly released by Neuroquery. Each sample in the dataset consists of a preprocessed pair of text and a corresponding brain functional activation map\cite{nelson2010role}. Note that, currently, only the titles, the most easily accessible content, are used as the foundational input to simulate the small-sample dataset. Also, relative to abstract and experimental design descriptions, titles contain fewer words such that the performance of the model can be pushed to the limit and a short phrase is more similar to the semantic query. Additionally, Neuroquery provides reported peak activation coordinates for each research article, assumed to be in the Montreal Neurological Institute 152 (MNI152)\cite{lancaster2007bias} brain 3D volumetric space. Following Neuroquery's preprocessing procedure, a Gaussian sphere with full width at half maximum (FWHM) of 9mm is placed at each peak activation coordinate, resulting in a brain activation map that serves as the target for prediction. The dataset is subsequently divided into a training set (8,076 samples), a validation set (2,092 samples), and a test set (2,092 samples) using a 6:2:2 ratio.


\subsection{Model}

Inspired by Text2Brain, Chat2Brain employs a similar architecture to map inputs in text modality to brain activation maps. As detailed in Fig.\ref{fig:enter-label}, it consists of a transformer-based text encoder, SciBERT, and a 3D CNN serving as the brain activation map generator. The text encoder optimizes the representation of the input and maps it to a 768-dimensional tensor, which is subsequently reshaped into a 4x5x4 3D volume with 64 channels. The brain activation map generator comprises three transposed 3D convolutional layers with channel sizes of 32, 16, and 8, respectively. Ultimately, it outputs brain activation maps with dimensions of 40x48x40. This architecture allows for the transformation of semantic representations into meaningful representations of brain activation patterns, facilitating the generation of accurate and informative brain activation maps based on the given inputs in text modality.

\begin{figure*}
    \centering
    \includegraphics[width=0.7\linewidth]{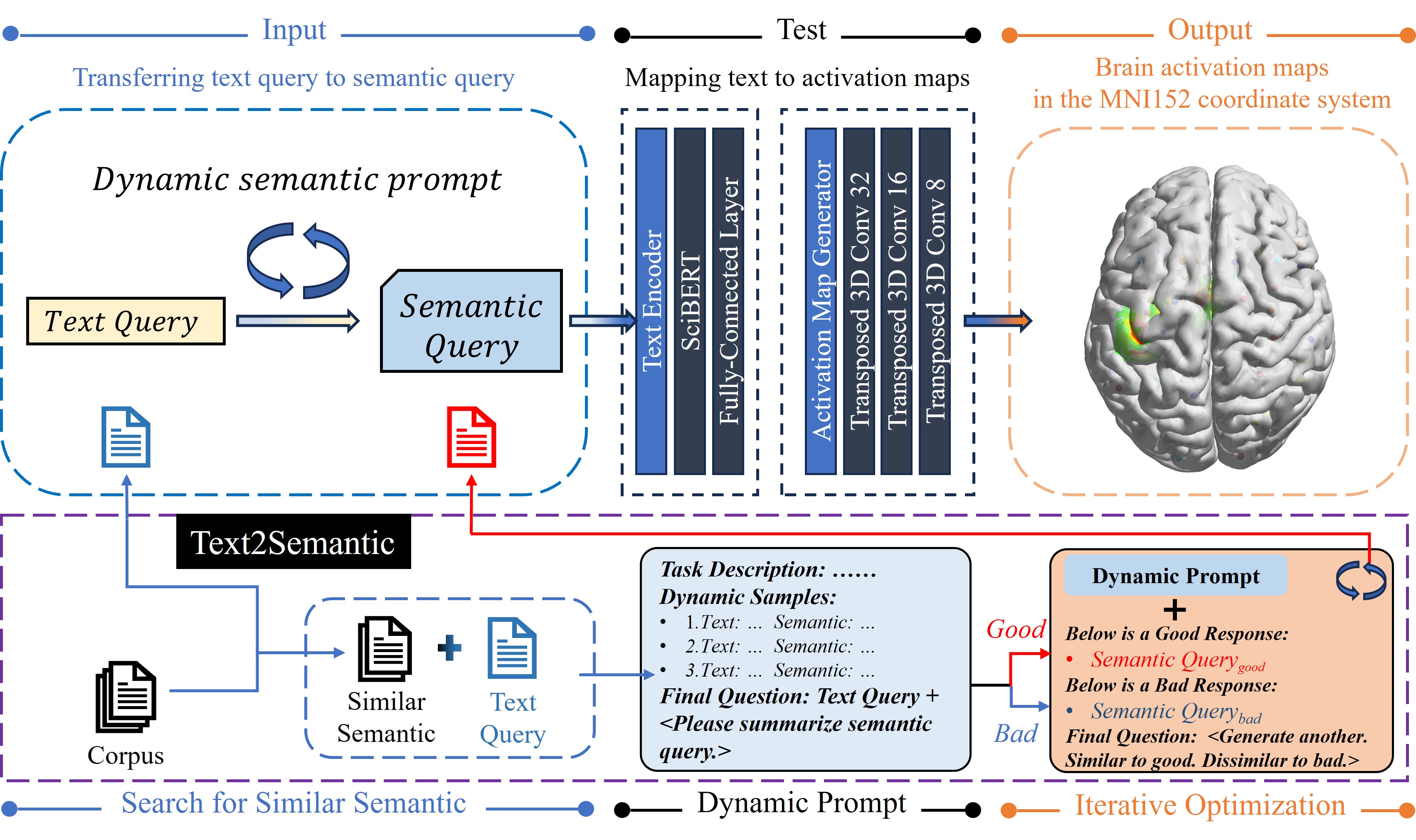}
    \caption{Flowchart of Chat2Brain during the testing phase.}
    \label{fig:enter-labe2}
\end{figure*}

\subsection{ChatAug}

During the training phase (in Fig.\ref{fig:enter-label}), ChatGPT which has accumulated a vast amount of knowledge across various domains including neuroscience, is capable of expanding the foundational text into a more diverse range of semantic information. This allows the model to learn more comprehensive semantic features, particularly when the dataset is limited in size and homogeneous in nature. In this experiment, for each sample, ChatGPT 3.5 is employed to augment the data with five types of semantic information: (1) new titles that are synonymous with the original title but exhibit significant differences, (2) new titles that are synonymous with the original title but exhibit minimal differences, (3) potential abstract content, (4) potential experimental design description, and (5) potential keywords. Particularly, the reason for selecting these five types of texts by augmentation is that they exhibit typical semantic information in neuroscience articles and vary in length, showcasing good diversity. For these reasons, the dataset, which initially has a limited number of samples and a narrow range of types, is transformed into a larger dataset with diverse types, thereby enhancing the generalizability of the model to handle a broader range of semantic query scenarios.

\begin{table}[htbp]
\centering
\caption{Accuracy of the method.}
\label{tab4}
\resizebox{0.5\textwidth}{!}
{
\begin{tabular}{l c c}
\noalign{\smallskip}
\toprule[2pt]
\noalign{\smallskip}
\makebox[0.2\textwidth][l]{Text/Metrics} &
\makebox[0.1\textwidth][c]{Semantics} &
\makebox[0.1\textwidth][c]{Detail}\\
\noalign{\smallskip}
\hline
\noalign{\smallskip}
title1 &100$\%$ &89.5$\%$\\
\noalign{\smallskip}
title2 &100$\%$ &94.7$\%$\\
\noalign{\smallskip}
abstract &94.7$\%$ &36.8$\%$\\
\noalign{\smallskip}
experiment &89.5$\%$ &15.8$\%$\\
\noalign{\smallskip}
keywords &84.2$\%$ &26.3$\%$\\
\noalign{\smallskip}
\bottomrule[2pt]
\noalign{\smallskip}
\end{tabular}
}
\vspace{-6pt}
\end{table}

Table \ref{tab4} presents the accuracy of randomly selected samples that have been augmented using ChatGPT compared with the real content in articles in terms of semantic and detail accuracy. Clearly, the augmented text demonstrates a high level of semantic accuracy and relatively lower detail accuracy. This aligns with cognitive expectations since it is difficult to augment the various text with highly consistent details based solely on the content of the title. However, Chat2Brain can extract the key semantic information from texts using Text2Semantics, where the detail is not very important.

\subsection{Text2Semantics}
Fig.\ref{fig:enter-labe2} illustrates the framework of Chat2Brain during testing. Although the model can cope with various semantic queries when the training phase is completed, the content of queries is not always accurate, concise, or even complete in practical application resulting in the wrong prediction. To avoid this situation, it is necessary to use ChatGPT to optimize queries and extract key semantic information before feeding it into the model during the test phase. This process is the transformation from text queries to semantic queries, called Text2Semantics.

Purple block in Fig.\ref{fig:enter-labe2} illustrates the flowchart of Text2Semantics, which is inspired by ImpressionGPT and can be decomposed into three sub-modules in processing order: Search for Similar Semantic, Dynamic Prompt, and Iterative Optimization. When Chat2Brain receives a text query, it first searches for similar samples in the corpus (training set) based on keywords (Search for Similar Semantic). These similar samples are engaged in research content that is closely related to the text query, sharing similar semantics. Subsequently, these similar samples are combined with the text query as dynamic prompts and fed into ChatGPT to summarize the key semantics of the text query as the initial semantic query (blue block in Dynamic Prompt). After generating the initial semantic query, an iterative optimization phase follows (orange block in Iterative Optimization). During each iteration, Chat2Brain assesses the similarity between the last semantic query and the similar samples in the corpus, determining whether they represent good (high similarity) or bad (low similarity) generations. Positive (good generations) or negative (bad generations) examples are then incorporated into the dynamic prompt to guide ChatGPT in generating new semantic queries for the next iteration. At the end of the iterations, the best-performing semantic query (highest similarity) is selected and mapped to brain activation maps in subsequent models. Thus, Text2Semantics serves as an intermediate bridge connecting text-to-semantics and semantics-to-brain activation map relationships, significantly enhancing the performance of Chat2Brain when confronted with complex and non-standard text queries.

\section{Experimental Setup}

\subsection{Settings}
During training, text-activation map pairs are randomly selected from the training set with equal probabilities. Each sample is fed into the model in the order of (1) the original title, (2) a title with significantly different synonymous words, (3) a title with minimally different synonymous words, (4) an abstract inferred by ChatGPT, (5) an experimental design description inferred by ChatGPT, (6) keywords inferred by ChatGPT, and (7) the original title. This approach encourages the model to generate outputs based on diverse semantic contexts, with varying text lengths and content, establishing a many-to-one mapping between semantics and brain activation maps.

The model is trained using the following parameter settings: (1) loss function: Mean Squared Error (MSE), (2) epochs: 2000, (3) optimizer: AdamW, and (4) learning rates: 1e-5 for the text encoder and 3e-2 for the brain activation map generator.

\begin{table*}[htbp]
\centering
\caption{Performance comparison of the Chat2Brain method.}
\label{tab1}
\resizebox{1.0\textwidth}{!}
{
\normalsize
\begin{tabular}{l c c c c c c c c c}
\noalign{\smallskip}
\toprule[2pt]
\noalign{\smallskip}
\multirow{2}{*}{\makebox[0.2\textwidth][l]{Text/Metrics}} &
\multicolumn{3}{c}{\makebox[0.1\textwidth][c]{Auc}} &
\multicolumn{3}{c}{\makebox[0.1\textwidth][c]{Dice}} &
\multicolumn{3}{c}{\makebox[0.1\textwidth][c]{mIou}}\\
\cmidrule(lr){2-4} \cmidrule(lr){5-7} \cmidrule(lr){8-10}
 & 
 \makebox[0.1\textwidth][c]{non-aug} &
 \makebox[0.1\textwidth][c]{aug} &
 \makebox[0.1\textwidth][c]{over} &
 \makebox[0.1\textwidth][c]{non-aug} &
 \makebox[0.1\textwidth][c]{aug} &
 \makebox[0.1\textwidth][c]{over} &
 \makebox[0.1\textwidth][c]{non-aug} &
 \makebox[0.1\textwidth][c]{aug} &
 \makebox[0.1\textwidth][c]{over} \\
\noalign{\smallskip}
\hline
\noalign{\smallskip}
title-100 &{0.599} &\textbf{0.631} &\textbf{+ 5.384$\textbf{\%}$} &{0.222} &\textbf{0.236} &\textbf{+ 6.285$\textbf{\%}$} &{0.131} &\textbf{0.140} &\textbf{+ 7.457$\textbf{\%}$}\\
\noalign{\smallskip}
title1-100 &{0.598} &\textbf{0.630} &\textbf{+ 5.307$\textbf{\%}$} &{0.222} &\textbf{0.236} &\textbf{+ 6.206$\textbf{\%}$} &{0.130} &\textbf{0.140} &\textbf{+ 7.358$\textbf{\%}$}\\
\noalign{\smallskip}
title2-100 &{0.598} &\textbf{0.630} &\textbf{+ 5.395$\textbf{\%}$} &{0.222} &\textbf{0.236} &\textbf{+ 6.288$\textbf{\%}$} &{0.130} &\textbf{0.140} &\textbf{+ 7.459$\textbf{\%}$}\\
\noalign{\smallskip}
keywords-100 &{0.592} &\textbf{0.638} &\textbf{+ 7.751$\textbf{\%}$} &{0.220} &\textbf{0.240} &\textbf{+ 9.026$\textbf{\%}$} &{0.129} &\textbf{0.142} &\textbf{+ 10.698$\textbf{\%}$}\\
\noalign{\smallskip}
abstract-100 &\textbf{0.640} &{0.639} &{- 0.133$\%$} &\textbf{0.241} &{0.240} &{- 0.579$\%$} &\textbf{0.144} &{0.143} &{- 0.584$\%$}\\
\noalign{\smallskip}
experiment-100 &{0.591} &\textbf{0.642} &\textcolor{red}{+ 8.725$\%$} &{0.219} &\textbf{0.241} &\textcolor{red}{+ 10.272$\%$} &{0.128} &\textbf{0.144} &\textcolor{red}{+ 12.190$\%$}\\
\noalign{\smallskip}
\midrule[2pt]
\noalign{\smallskip}
\multirow{2}{*}{\makebox[0.2\textwidth][l]{Text/Metrics}} &
\multicolumn{3}{c}{\makebox[0.1\textwidth][c]{Auc}} &
\multicolumn{3}{c}{\makebox[0.1\textwidth][c]{Dice}} &
\multicolumn{3}{c}{\makebox[0.1\textwidth][c]{mIou}}\\
\cmidrule(lr){2-4} \cmidrule(lr){5-7} \cmidrule(lr){8-10}
 & 
 \makebox[0.1\textwidth][c]{non-chat} &
 \makebox[0.1\textwidth][c]{chat} &
 \makebox[0.1\textwidth][c]{over} &
 \makebox[0.1\textwidth][c]{non-chat} &
 \makebox[0.1\textwidth][c]{chat} &
 \makebox[0.1\textwidth][c]{over} &
 \makebox[0.1\textwidth][c]{non-chat} &
 \makebox[0.1\textwidth][c]{chat} &
 \makebox[0.1\textwidth][c]{over} \\
\noalign{\smallskip}
\hline
\noalign{\smallskip}
non-aug-100 &{0.599} &\textbf{0.601} &\textcolor{red}{+ 0.282$\%$} &{0.222} &\textbf{0.223} &\textbf{+ 0.254$\textbf{\%}$} &{0.131} &\textbf{0.132} &\textbf{+ 0.297$\textbf{\%}$}\\
\noalign{\smallskip}
non-aug-90 &\textbf{0.802} &{0.798} &{- 0.551$\%$} &{0.065} &\textbf{0.066} &\textbf{+ 1.938$\textbf{\%}$} &{0.034} &\textbf{0.035} &\textbf{+ 2.054$\textbf{\%}$}\\
\noalign{\smallskip}
non-aug-80 &\textbf{0.713} &{0.702} &{- 1.543$\%$} &{0.056} &\textbf{0.058} &\textbf{+ 2.200$\textbf{\%}$} &{0.030} &\textbf{0.031} &\textbf{+ 2.287$\textbf{\%}$}\\
\noalign{\smallskip}
non-aug-70 &\textbf{0.637} &{0.626} &{- 1.772$\%$} &{0.048} &\textbf{0.049} &\textbf{+ 1.212$\textbf{\%}$} &{0.025} &\textbf{0.026} &\textbf{+ 1.240$\textbf{\%}$}\\
\noalign{\smallskip}
non-aug-60 &\textbf{0.587} &{0.578} &{- 1.553$\%$} &{0.033} &\textbf{0.034} &\textbf{+ 0.530$\textbf{\%}$} &{0.017} &\textbf{0.018} &\textbf{+ 0.495$\textbf{\%}$}\\
\noalign{\smallskip}
non-aug-50 &\textbf{0.555} &{0.549} &{- 1.063$\%$} &\textbf{0.027} &{0.026} &{- 1.667$\%$} &\textbf{0.015} &{0.014} &{- 1.704$\%$}\\
\noalign{\smallskip}
non-aug-40 &\textbf{0.535} &{0.532} &{- 0.472$\%$} &\textbf{0.015} &{0.014} &{- 2.064$\%$} &\textbf{0.009} &{0.008} &{- 2.225$\%$}\\
\noalign{\smallskip}
non-aug-30 &\textbf{0.519} &{0.518} &{- 0.210$\%$} &\textbf{0.011} &{0.010} &{- 3.701$\%$} &\textbf{0.006} &{0.005} &{- 3.812$\%$}\\
\noalign{\smallskip}
non-aug-20 &{0.509} &\textbf{0.510} &\textbf{+ 0.227$\textbf{\%}$} &{0.007} &\textbf{0.008} &\textcolor{red}{+ 4.674$\%$} &{0.004} &\textbf{0.005} &\textcolor{red}{+ 4.730$\%$}\\
\noalign{\smallskip}
non-aug-10 &{0.505} &\textbf{0.506} &\textbf{+ 0.069$\textbf{\%}$} &{0.004} &\textbf{0.005} &\textbf{+ 3.070$\textbf{\%}$} &{0.002} &\textbf{0.003} &\textbf{+ 4.286$\textbf{\%}$}\\
\noalign{\smallskip}
\noalign{\smallskip}
\hline
\noalign{\smallskip}
\noalign{\smallskip}
aug-100 &{0.631} &\textbf{0.633} &\textcolor{red}{+ 0.296$\%$} &{2.361e-1} &\textbf{2.370e-1} &\textbf{+ 0.377$\textbf{\%}$} &{1.403e-1} &\textbf{1.409e-1} &\textbf{+ 0.450$\textbf{\%}$}\\
\noalign{\smallskip}
aug-90 &\textbf{0.796} &{0.795} &{- 0.071$\%$} &\textbf{3.922e-1} &{3.908e-1} &{- 0.371$\%$} &\textbf{2.026e-1} &{2.018e-1} &{- 0.385$\%$}\\
\noalign{\smallskip}
aug-80 &{0.795} &\textbf{0.796} &\textbf{+ 0.047$\textbf{\%}$} &\textbf{2.747e-1} &{2.733e-1} &{- 0.513$\%$} &\textbf{1.404e-2} &{1.396e-2} &{- 0.527$\%$}\\
\noalign{\smallskip}
aug-70 &{0.754} &\textbf{0.755} &\textbf{+ 0.182$\textbf{\%}$} &\textbf{1.984e-2} &{1.973e-2} &{- 0.552$\%$} &\textbf{1.007e-2} &{1.001e-2} &{- 0.566$\%$}\\
\noalign{\smallskip}
aug-60 &{0.659} &\textbf{0.660} &\textbf{+ 0.067$\textbf{\%}$} &\textbf{1.255e-2} &{1.246e-2} &{- 0.697$\%$} &\textbf{6.347e-3} &{6.303e-3} &{- 0.705$\%$}\\
\noalign{\smallskip}
aug-50 &{0.581} &\textbf{0.582} &\textbf{+ 0.049$\textbf{\%}$} &\textbf{1.006e-2} &{9.951e-3} &{- 1.126$\%$} &\textbf{5.088e-3} &{5.029e-3} &{- 1.144$\%$}\\
\noalign{\smallskip}
aug-40 &\textbf{0.542} &{0.541} &{- 0.084$\%$} &\textbf{4.692e-3} &{4.588e-3} &{- 2.224$\%$} &\textbf{2.367e-3} &{2.314e-3} &{- 2.246$\%$}\\
\noalign{\smallskip}
aug-30 &{0.520} &\textbf{0.521} &\textbf{+ 0.039$\textbf{\%}$} &\textbf{3.411e-3} &{3.354e-3} &{- 1.672$\%$} &\textbf{1.724e-3} &{1.695e-3} &{- 1.678$\%$}\\
\noalign{\smallskip}
aug-20 &{0.507} &\textbf{0.508} &\textbf{+ 0.029$\textbf{\%}$} &{2.139e-3} &\textbf{2.175e-3} &\textbf{+ 1.653$\textbf{\%}$} &{1.087e-3} &\textbf{1.105e-3} &\textbf{+ 1.618$\textbf{\%}$}\\
\noalign{\smallskip}
aug-10 &{0.501} &\textbf{0.502} &\textbf{+ 0.026$\textbf{\%}$} &{1.275e-3} &\textbf{1.340e-3} &\textcolor{red}{+ 5.093$\%$} &{6.658e-4} &\textbf{6.996e-4} &\textcolor{red}{+ 5.064$\%$}\\
\noalign{\smallskip}
\bottomrule[2pt]
\noalign{\smallskip}
\end{tabular}
}
\vspace{-6pt}
\end{table*}


\subsection{Task}
The experiment is conducted with two different query environments: standard and non-standard, aiming to simulate real-world usage scenarios. In the standard query environment, the text query is complete, minimally redundant, and semantically accurate. However, in a real application (the non-standard query environment), the text query may suffer from issues such as text incompleteness, high redundancy, and semantic inaccuracies. To simulate the non-standard query environment, random masking is applied to the input to disrupt its semantics. In both query environments, the experiment compared the performance before and after transforming the text query into the semantic query. Additionally, the experiment evaluates the predictive performance of the model under various text contexts before and after applying ChatAUG for data augmentation.

\subsection{Baselines}
Neuroquery and Text2Brain are chosen as the baseline. Chat2Brain shares the same model architecture as Text2Brain. The only difference is that Chat2Brain does not involve a large language model, and its input consists of regular text queries.

\subsection{Evaluation Metrics}
In this experiment, Area Under the Curve (AUC)\cite{myerson2001area}, Dice Score\cite{dice1945measures}, and mean Intersection over Union (mIoU) are selected as metrics to measure the similarity between the predicted and target brain activation maps at different levels of detail. The evaluation process is the same as that of Text2Brain. For both the predicted and target brain activation maps, the same proportion of the most relevant activated voxels is retained (ranging from 100$\%$ retention to 10$\%$ retention).

\begin{table*}[htbp]
\centering
\caption{Performance comparison of the Chat2Brain method.}
\label{tab2}
\resizebox{1.0\textwidth}{!}
{
\normalsize
\begin{tabular}{l c c c c c c c c c}
\noalign{\smallskip}
\toprule[2pt]
\noalign{\smallskip}
\multirow{2}{*}{\makebox[0.2\textwidth][l]{Method/Metrics}} &
\multicolumn{3}{c}{\makebox[0.1\textwidth][c]{Auc}} &
\multicolumn{3}{c}{\makebox[0.1\textwidth][c]{Dice}} &
\multicolumn{3}{c}{\makebox[0.1\textwidth][c]{mIou}}\\
\cmidrule(lr){2-4} \cmidrule(lr){5-7} \cmidrule(lr){8-10}
 & 
 \makebox[0.1\textwidth][c]{non-chat} &
 \makebox[0.1\textwidth][c]{chat} &
 \makebox[0.1\textwidth][c]{over} &
 \makebox[0.1\textwidth][c]{non-chat} &
 \makebox[0.1\textwidth][c]{chat} &
 \makebox[0.1\textwidth][c]{over} &
 \makebox[0.1\textwidth][c]{non-chat} &
 \makebox[0.1\textwidth][c]{chat} &
 \makebox[0.1\textwidth][c]{over} \\
\noalign{\smallskip}
\hline
\noalign{\smallskip}
non-aug-100 &\textbf{0.601} &{0.599} &{- 0.524$\%$} &\textbf{0.223} &{0.222} &{- 0.600$\%$} &\textbf{0.131} &{0.130} &{- 0.725$\%$}\\
\noalign{\smallskip}
non-aug-90 &{0.801} &\textbf{0.802} &\textbf{+ 0.087$\textbf{\%}$} &{0.058} &\textbf{0.065} &\textbf{+ 10.078$\textbf{\%}$} &{0.031} &\textbf{0.034} &\textbf{+ 10.844$\textbf{\%}$}\\
\noalign{\smallskip}
non-aug-80 &{0.705} &\textbf{0.713} &\textbf{+ 1.102$\textbf{\%}$} &{0.047} &\textbf{0.056} &\textbf{+ 19.373$\textbf{\%}$} &{0.025} &\textbf{0.030} &\textbf{+ 20.782$\textbf{\%}$}\\
\noalign{\smallskip}
non-aug-70 &{0.623} &\textbf{0.637} &\textbf{+ 2.180$\textbf{\%}$} &{0.038} &\textbf{0.048} &\textbf{+ 25.467$\textbf{\%}$} &{0.019} &\textbf{0.025} &\textbf{+ 27.060$\textbf{\%}$}\\
\noalign{\smallskip}
non-aug-60 &{0.573} &\textbf{0.587} &\textcolor{red}{+ 2.408$\%$} &{0.025} &\textbf{0.033} &\textbf{+ 30.029$\textbf{\%}$} &{0.013} &\textbf{0.017} &\textbf{+ 31.592$\textbf{\%}$}\\
\noalign{\smallskip}
non-aug-50 &{0.544} &\textbf{0.555} &\textbf{+ 2.034$\textbf{\%}$} &{0.020} &\textbf{0.027} &\textbf{+ 32.168$\textbf{\%}$} &{0.010} &\textbf{0.014} &\textbf{+ 33.586$\textbf{\%}$}\\
\noalign{\smallskip}
non-aug-40 &{0.528} &\textbf{0.535} &\textbf{+ 1.325$\textbf{\%}$} &{0.010} &\textbf{0.015} &\textbf{+ 37.012$\textbf{\%}$} &{0.006} &\textbf{0.008} &\textbf{+ 39.200$\textbf{\%}$}\\
\noalign{\smallskip}
non-aug-30 &{0.516} &\textbf{0.519} &\textbf{+ 0.741$\textbf{\%}$} &{0.008} &\textbf{0.010} &\textbf{+ 30.711$\textbf{\%}$} &{0.004} &\textbf{0.006} &\textbf{+ 32.022$\textbf{\%}$}\\
\noalign{\smallskip}
non-aug-20 &{0.508} &\textbf{0.509} &\textbf{+ 0.189$\textbf{\%}$} &{0.006} &\textbf{0.007} &\textbf{+ 21.982$\textbf{\%}$} &{0.003} &\textbf{0.004} &\textbf{+ 23.263$\textbf{\%}$}\\
\noalign{\smallskip}
non-aug-10 &{0.503} &\textbf{0.505} &\textbf{+ 0.344$\textbf{\%}$} &{0.003} &\textbf{0.005} &\textcolor{red}{+ 55.934$\%$} &{0.002} &\textbf{0.003} &\textcolor{red}{+ 56.829$\%$}\\
\noalign{\smallskip}
\noalign{\smallskip}
\hline
\noalign{\smallskip}
\noalign{\smallskip}
aug-100 &\textbf{0.646} &{0.631} &{- 2.486$\%$} &\textbf{2.439e-1} &{2.361e-1} &{- 3.199$\%$} &\textbf{1.458e-1} &{1.403e-1} &{- 3.772$\%$}\\
\noalign{\smallskip}
aug-90 &{0.790} &\textbf{0.796} &\textcolor{red}{+ 0.765$\%$} &{3.822e-2} &\textbf{3.922e-2} &\textbf{+ 2.608$\textbf{\%}$} &{1.973e-2} &\textbf{2.026e-2} &\textbf{+ 2.690$\textbf{\%}$}\\
\noalign{\smallskip}
aug-80 &{0.805} &\textbf{0.807} &\textbf{+ 0.253$\textbf{\%}$} &{2.662e-2} &\textbf{2.747e-2} &\textbf{+ 3.220$\textbf{\%}$} &{1.259e-2} &\textbf{1.404e-2} &\textbf{+ 3.288$\textbf{\%}$}\\
\noalign{\smallskip}
aug-70 &\textbf{0.760} &{0.754} &{- 0.726$\%$} &{1.904e-2} &\textbf{1.984e-2} &\textbf{+ 4.196$\textbf{\%}$} &{9.664e-3} &\textbf{1.008e-2} &\textbf{+ 4.263$\textbf{\%}$}\\
\noalign{\smallskip}
aug-60 &\textbf{0.666} &{0.659} &{- 1.297$\%$} &{1.190e-2} &\textbf{1.255e-2} &\textbf{+ 5.445$\textbf{\%}$} &{6.015e-3} &\textbf{6.348e-3} &\textbf{+ 5.519$\textbf{\%}$}\\
\noalign{\smallskip}
aug-50 &\textbf{0.587} &{0.581} &{- 0.898$\%$} &{9.433e-3} &\textbf{1.006e-2} &\textbf{+ 6.688$\textbf{\%}$} &{4.764e-3} &\textbf{5.088e-3} &\textbf{+ 6.799$\textbf{\%}$}\\
\noalign{\smallskip}
aug-40 &\textbf{0.542} &{0.541} &{- 0.353$\%$} &{4.289e-3} &\textbf{4.692e-3} &\textbf{+ 9.388$\textbf{\%}$} &{2.161e-3} &\textbf{2.367e-3} &\textbf{+ 9.534$\textbf{\%}$}\\
\noalign{\smallskip}
aug-30 &\textbf{0.520} &{0.519} &{- 0.135$\%$} &{3.086e-3} &\textbf{3.411e-3} &\textcolor{red}{+ 10.527$\%$} &{1.557e-3} &\textbf{1.724e-3} &\textcolor{red}{+ 10.712$\%$}\\
\noalign{\smallskip}
aug-20 &\textbf{0.507} &{0.506} &{- 0.210$\%$} &{2.043e-3} &\textbf{2.140e-3} &\textbf{+ 4.727$\textbf{\%}$} &{1.035e-3} &\textbf{1.087e-3} &\textbf{+ 5.011$\textbf{\%}$}\\
\noalign{\smallskip}
aug-10 &\textbf{0.502} &{0.501} &{- 0.078$\%$} &{1.262e-3} &\textbf{1.276e-3} &\textbf{+ 1.029$\textbf{\%}$} &{6.534e-4} &\textbf{6.669e-4} &\textbf{+ 1.911$\textbf{\%}$}\\
\noalign{\smallskip}
\bottomrule[2pt]
\noalign{\smallskip}
\end{tabular}
}
\vspace{-6pt}
\end{table*}


\section{Results}

\subsection{Task in the standard query environment}
Table \ref{tab1} presents the experimental results in the standard query environment. In the table, "title," "title1," "title2," "keywords," "abstract," and "experiment" represent the original title, two titles with synonymous and paraphrased words, keywords, abstract, and experimental tasks inferred by ChatGPT. "non-aug" and "aug" denote the absence or presence of data augmentation using ChatAUG. $+ X\%$ denotes an $X\%$ improvement compared to non-augmentation when using augmentation. $- X\%$ denotes an $X\%$ decrease compared to non-augmentation when using augmentation. "non-chat" and "chat" indicate the absence or utilization of Text2Semantics to transform the text queries into semantic queries. $+ X\%$ denotes an $X\%$ improvement compared to text queries when using semantic queries. $- X\%$ denotes an $X\%$ decrease compared to text queries when using semantic queries.
 The digit $k$ in "aug-$k$" denotes that the top $k\%$ of the predicted activation voxels are preserved.

Observing the results in Table \ref{tab1}, we find that: (1) In most cases, Chat2Brain with data augmentation performs better in handling diverse types of input queries compared to Text2Brain. It shows the highest improvements in AUC, Dice, and mIOU, reaching up to 8.725$\%$, 10.272$\%$, and 12.190$\%$, respectively (marked in red color). (2) When faced with excessively long queries, such as abstracts in the first section, Chat2Brain performs relatively worse, but the differences in all three metrics compared to Text2Brain do not exceed 0.584$\%$. (3) Without data augmentation, semantic queries do not exhibit significant advantages over text queries. However, when only a smaller number of the most relevant activated voxels are retained (non-aug-/aug-$k$, $k<30$, in general), semantic queries slightly outperform text queries. (4) With data augmentation, semantic queries have a certain advantage over text queries (aug- rows plus chat columns), especially when the most activated voxels are retained. The highest improvements in the three metrics can reach 0.029$\%$, 5.093$\%$, and 5.064$\%$, respectively (marked in red color).

The reasons behind (3) and (4) may lie in the fact that in the standard query environment, text queries themselves already possess comprehensive and refined semantics. Therefore, transforming them into semantic queries does not bring significant enhancements. On the other hand, the model with data augmentation sacrifices some precision when only the most activated voxels are retained, in exchange for better generality. As a result, it becomes more sensitive to subtle semantic variations, leading to superior performance in semantic queries.

In conclusion, in the standard query environment, Chat2Brain demonstrates better generalizability and comparable predictive accuracy than Text2Brain.

\subsection{Task in the non-standard query environment}
Table \ref{tab2} presents the experimental results in the non-standard query environment.

Observing the results in Table \ref{tab2}, the following findings can be noted: (1) Without data augmentation, semantic queries have a significant advantage over text queries, showing the highest improvements in the three metrics, reaching up to 2.408$\%$, 55.934$\%$, and 56.829$\%$, respectively (marked in red color). (2) Even with data augmentation, semantic queries still exhibit better performance, with the highest improvements in the three metrics reaching up to 0.765$\%$, 10.527$\%$, and 10.712$\%$, respectively (marked in red color).

The reasons behind the differences in (1) and (2) are attributed to that the model without data augmentation, struggles to handle diverse and complex queries, leading to a significant decrease in predictive accuracy. Thus, leveraging Text2Semantics to refine and restore the semantics results in a significant leap in performance. On the other hand, the model with data augmentation, already possesses the capability to handle complex queries, resulting in a slight decrease in accuracy, and therefore, the improvements after semantic restoration are also comparatively lower. At the same time, we notice that the augmented models, while retaining fewer voxels, have an extremely low negative impact on AUC when using semantic queries. However, compared to the significant improvement in Dice and MIoU, this impact can be considered negligible.

\subsection{Comparison with SOTAs}
Results in Table \ref{tab3} demonstrate that Chat2Brain outperforms Text2Brain and NeuroQuery, both in standard and non-standard query environments, with the most relevant voxels retained in the prediction results (10$\%$ or 20$\%$).

\begin{table}[htbp]
\centering
\caption{Performance comparison of the Chat2Brain method.}
\label{tab3}
\resizebox{0.5\textwidth}{!}
{
\normalsize
\begin{tabular}{l c c c c}
\noalign{\smallskip}
\toprule[2pt]
\noalign{\smallskip}
\multirow{2}{*}{\makebox[0.2\textwidth][l]{Method/Metrics}} &
\multicolumn{2}{c}{\makebox[0.1\textwidth][c]{Standard}} &
\multicolumn{2}{c}{\makebox[0.1\textwidth][c]{Non-Standard}}\\
\cmidrule(lr){2-3} \cmidrule(lr){4-5}
 & 
 \makebox[0.1\textwidth][c]{Dice} &
 \makebox[0.1\textwidth][c]{mIou} &
 \makebox[0.1\textwidth][c]{Dice} &
 \makebox[0.1\textwidth][c]{mIou} \\
\noalign{\smallskip}
\hline
\noalign{\smallskip}
neuroquery-20 &{6.442e-4} &{3.224e-4} &{6.414e-4} &{3.210e-4}\\
\noalign{\smallskip}
text2brain-20 &{2.139e-3} &{1.087e-3} &{2.043e-3} &{1.035e-3}\\
\noalign{\smallskip}
chat2brain-20 &\textbf{2.175e-3} &\textbf{1.105e-3} &\textbf{2.140e-3} &\textbf{1.087e-3}\\
\noalign{\smallskip}
\hline
\noalign{\smallskip}
neuroquery-10 &{4.570e-4} &{2.287e-4} &{4.442e-4} &{2.222e-4}\\
\noalign{\smallskip}
text2brain-10 &{1.275e-3} &{6.658e-4} &{1.262e-3} &{6.534e-4}\\
\noalign{\smallskip}
chat2brain-10 &\textbf{1.341e-3} &\textbf{6.996e-4} &\textbf{1.276e-3} &\textbf{6.665e-4}\\
\noalign{\smallskip}
\bottomrule[2pt]
\noalign{\smallskip}
\end{tabular}
}
\vspace{-6pt}
\end{table}

\begin{figure}
    \centering
    \includegraphics[width=1\linewidth]{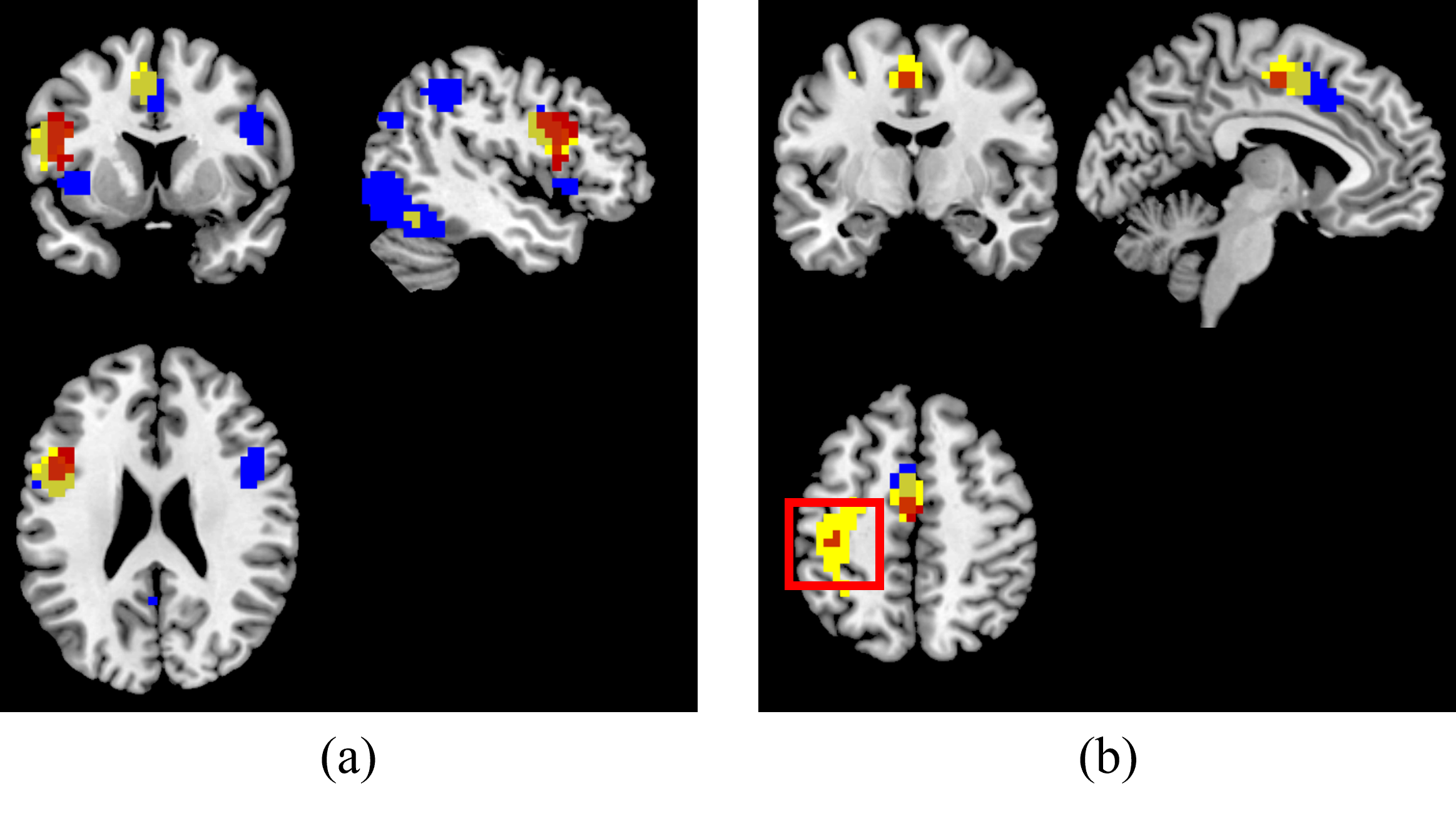}
    \caption{Results of Chat2Brain and Text2Brain in two sutdies under non-standard environment. The red regions are the ground truth, the blue regions are the predicted results from Text2Brain, and the yellow is the predicted results from Chat2Brain. Interpretation of the red box is referred to the text.}
    \label{fig:enter-labe4}
\end{figure}

While Text2Brain can synthesize meaningful neural activation patterns from free-form textual descriptions, some activation patterns learned from certain published studies are found at incorrect positions. Compared to Text2Brain, our model can learn deeper and further from input queries, then generate more accurate activation patterns. Fig.\ref{fig:enter-labe4} shows two such examples. As reported in \cite{matsuo2010neural}, researchers studied activation patterns when subjects speak the Japanese character Kanji, increased activation patterns are shown in posterior regions of the left, middle, and inferior frontal gyri, which is viewed as a part of Broca Area for language processing (red regions in Fig.\ref{fig:enter-labe4}(a)). Text2brain predicts activation patterns in the right hemisphere (blue regions), but our model correctly predicts these patterns near the left inferior frontal gyri (yellow regions). In another study where brain patterns with different handgrip types are compared\cite{king2014handyman}, our method could not only learn more precise activation patterns in the pre-central gyrus from another study but also find activation patterns in the middle post central gyrus near Brodmann Area 4 compared to Text2Brain (red-boxed region in Fig.\ref{fig:enter-labe4}(b)). In conclusion, our method can predict more precise activation patterns by reorganizing textual descriptions and learning semantic queries.

Taking into account the results in the standard and non-standard environment, it can be concluded that Chat2Brain establishes a two-stage mapping model with high generalizability, capable of handling complex and challenging query environments.


\section{Conclusion}
In this work, we proposed a Chat2Brain model that combines Text2Brain with ChatGPT to generate brain activation maps from semantic queries. By introducing ChatGPT for data augmentation using ChatAUG and semantic extraction using Text2Semantics during both training and testing, we transformed conventional text queries into more informative semantic queries. This established a model, Chat2Brain, with improved generalizability and the ability to handle complex query environments that may be encountered in practical applications. Chat2Brain not only captures rich relationships between different semantics but also leverages the accumulated knowledge associations of large language models to restore, refine, and expand these semantic relationships. It addresses the limitations of previous methods such as Text2Brain, Neuroquery, and Neurosynth in dealing with complex queries in real-world scenarios. Additionally, it bridges the gap between large language models and neuroscience through the modality of text, marking the first step in the development of large models in the field of neuroscience. This will contribute to further advancements and explorations. In the future, we will strive to develop large models specifically tailored to the field of neuroscience, linking the internal activations of the models with brain activations to explore more possibilities in neuroscience.

\bibliography{sample}

\vspace{12pt}

\end{document}